\pdfoutput=1
\documentclass[twoside,11pt]{article}
\usepackage{multirow}
\usepackage{blindtext}
\usepackage{listings}

\usepackage{jmlr2e}

\usepackage{lastpage}

\ShortHeadings{A Preprint}{KerasCV and KerasNLP: Vision and Language Power-Ups}
\firstpageno{1}

\begin{document}

\title{KerasCV and KerasNLP: Vision and Language Power-Ups}

    \author{\sc Lead Authors\\
        \name Matthew Watson,
        Divyashree Shivakumar Sreepathihalli,
        Fran\c{c}ois Chollet\\
        Martin G\"orner,
        Kiranbir Sodhia,
        Ramesh Sampath,
        Tirth Patel\\
        Haifeng Jin,
        Neel Kovelamudi,
        Gabriel Rasskin,
        Samaneh Saadat\\
        Luke Wood,
        Chen Qian,
        Jonathan Bischof,
        Ian Stenbit
        \AND
        \small\sc
        \{mattdangerw,
        divyasreepat,
        fchollet,
        mgorner,
        ksodhia\}@google.com\\
        \{rameshsampath,
        tirthp,
        haifengj,
        nkovela,
        grasskin,
        ssaadat\}@google.com\\
        \{lukewoodcs,
        qianchen94era,
        jbischof1\}@gmail.com,
        ian@stenbit.com
        \AND
        \addr{Keras Team, Google LLC, USA}
        \AND
        \AND
        \sc Community Contributors\\
        \name Abheesht Sharma, Anshuman Mishra
        \AND
        \small\sc
        \{sharmabhee,shivanshuman021\}@gmail.com
       }

\editor{Enrica Filippi, Robert Hundt}

\maketitle

\begin{abstract}
We present the Keras domain packages KerasCV and KerasNLP, extensions of the Keras API for Computer Vision and Natural Language Processing workflows, capable of running on either JAX, TensorFlow, or PyTorch. These domain packages are designed to enable fast experimentation, with a focus on ease-of-use and performance. We adopt a modular, layered design: at the library's lowest level of abstraction, we provide building blocks for creating models and data preprocessing pipelines, and at the library's highest level of abstraction, we provide pretrained ``task" models for popular architectures such as Stable Diffusion, YOLOv8, GPT2, BERT, Mistral, CLIP, Gemma, T5, etc. Task models have built-in preprocessing, pretrained weights, and can be fine-tuned on raw inputs. To enable efficient training, we support XLA compilation for all models, and run all preprocessing via a compiled graph of TensorFlow operations using the \verb|tf.data| API. The libraries are fully open-source (Apache 2.0 license) and available on GitHub.
\end{abstract}

\begin{keywords}
  KerasCV, KerasNLP, Keras multibackend, Deep learning, Generative AI
\end{keywords}

\section{Introduction}

Keras~\citep{chollet2015keras} is among the most widely used tools for machine learning today\footnote{\url{https://survey.stackoverflow.co/2022/}}. The Keras library acts as a high-level abstraction for machine learning models and layers, and seeks to be accessible to a broad group of machine learning researchers and practitioners by focusing on rapid experimentation and progressive disclosure of complexity.

Notably, recent developments in Computer Vision (CV) and Natural Language Processing (NLP) have created new challenges for practitioners. The most obvious is the shift towards larger and larger models trained on self-supervised tasks. Pretraining a state of the art model is now cost-prohibitive for many researchers and practitioners, in particular in NLP. Access to open-source model architectures with pretrained weights is imperative for a large amount of CV and NLP work.

Additionally, pairing efficient preprocessing and metrics computation for modern models has become more difficult, with a proliferation of disparate techniques, backends, and licenses. An ML researcher or practitioner today must select among a range of auto-differentiation frameworks such as JAX, TensorFlow, and PyTorch, and even within each framework, they are often forced to stay within a specific modeling library for cross-compatibility of components.

Further, improving the train-time performance of models on NLP problems presents additional hurdles. The XLA compiler~\citep{50530} offers dramatic speedups for many model architectures, but adds complex restrictions on the shape and flow of tensor operations. The TensorFlow-based \verb|tf.data|~\citep{DBLP:journals/corr/abs-2101-12127} and \verb|tf.text| APIs provide a scalable, dynamic, and multi-process approach for preprocessing, but many common text operations do not easily compile to a TensorFlow graph.

Aiming to reduce these framework barriers for CV and NLP practitioners and researchers, we present KerasCV and KerasNLP, extensions of the Keras API for computer vision and natural language processing workflows. These packages expand upon the modular approach of Keras, adding pretrained backbone models, easy-to-use domain-specific losses and metrics, out-of-the-box support for XLA~\citep{50530} compilation and data and model parallelism. Because these domain packages are written on top of Keras 3, all of their modeling components natively support JAX~\citep{jax2018github}, TensorFlow~\citep{tensorflow2015-whitepaper}, and PyTorch~\citep{NEURIPS2019_9015}, and can be freely used in framework-native workflows that do not otherwise involve any Keras components.

\section{The Keras Domain Packages API}

We adopt a layered approach to API design. Our library has three primary levels of abstraction:

\begin{itemize}
  \item \textbf{Foundational Components}: A collection of composable modules for building and training preprocessing pipelines, models, and evaluation logic. These are pure Keras 3 components which can be used outside of the Keras Domain Packages ecosystem.
  \item \textbf{Pretrained Backbones}: A collection of pretrained model backbones for fine-tuning. For NLP models, matching tokenizers can be created alongside backbones.
  \item \textbf{Task Models}: A collection of end-to-end models specialized for a specific task, e.g. text generation in NLP or object detection in CV. These task models combine the preprocessing and modeling modules from the lower API levels to create a unified training and inference interface that can operate directly on plain text or image input. Task models aim to allow fine-tuning with zero configuration for common use cases.
\end{itemize}

Each additional API layer is built on top of the previous one. Modules from each level can be mixed and matched in usage, for example, extending a pretrained backbone with foundational preprocessing modules to pack input sequences or perform data augmentation.

Any KerasCV and KerasNLP model can be instantiated as a PyTorch
\verb|torch.nn.Module|, a TensorFlow \verb|tf.Module|, or as a stateless JAX function. This means that the models can be used with PyTorch ecosystem packages, with the full range of TensorFlow deployment and production tools (such as TF-Serving, TF.js and TFLite), and with JAX large-scale TPU training infrastructure.

\section{Training, Serving, and Deployment}

KerasCV and KerasNLP offer large vision and language models. State of the art models are expected to continually increase in size in the future. To address these problems, KerasCV and KerasNLP are compatible with the Keras Unified Distribution API. This API enables both model parallelism and data parallelism across all Keras backends. The API maintains a clear separation between the model definition, training logic, and sharding configuration. As a result, models within KerasCV and KerasNLP can be written as if they were intended to run on a single device. Later, specific sharding configurations can be added to these models when it's time to train them.\footnote{\url{https://keras.io/api/distribution/}}

\section{Pretrained models on Kaggle Models}
All pretrained models of KerasCV and KerasNLP are published on Kaggle Models \footnote{\raggedright{\url{https://www.kaggle.com/organizations/keras/models}}}. Importantly, these models are also available on Kaggle competition notebooks in Internet-off mode.

\begin{table}[ht]
\scriptsize

\begin{tabular}{c|cc|cc|cc|cc|cc|}
\cline{1-9} 
 \multicolumn{1}{|l|}{}            & \multicolumn{2}{c|}{SAM} & \multicolumn{2}{c|}{Gemma}  & \multicolumn{2}{c|}{BERT}  & \multicolumn{2}{c|}{Mistral} \\ 
\multicolumn{1}{|l|}{}            & \multicolumn{1}{c|}{train} & predict & \multicolumn{1}{c|}{train} & predict & \multicolumn{1}{l|}{train} & predict & \multicolumn{1}{c|}{train} & predict \\ \hline
\multicolumn{1}{|l|}{Batch Size} & \multicolumn{1}{l|}{1} & \multicolumn{1}{l|}{7}  & \multicolumn{1}{l|}{8} & \multicolumn{1}{l|}{32} & \multicolumn{1}{l|}{54} & \multicolumn{1}{l|}{531}  & \multicolumn{1}{l|}{8} & \multicolumn{1}{l|}{32} \\ \hline
\multicolumn{1}{|l|}{Keras 2 (TF)}  & \multicolumn{1}{l|}{386.93} & \multicolumn{1}{l|}{3,187.09} & \multicolumn{1}{l|}{NA} & \multicolumn{1}{l|}{NA} & \multicolumn{1}{l|}{841.84} & \multicolumn{1}{l|}{965.21}  &  \multicolumn{1}{l|}{NA} & \multicolumn{1}{l|}{NA}\\ \hline

\multicolumn{1}{|l|}{Keras 3 (TF)} & \multicolumn{1}{l|}{\textbf{355.25}} &  \multicolumn{1}{l|}{762.67} & \multicolumn{1}{l|}{\textbf{232.52}} & \multicolumn{1}{l|}{1,134.91}  & \multicolumn{1}{l|}{\textbf{404.17}} & \multicolumn{1}{l|}{962.11}   & \multicolumn{1}{l|}{\textbf{185.92}} & \multicolumn{1}{l|}{966.06} \\ \hline

\multicolumn{1}{|l|}{Keras 3 (JAX)}  & \multicolumn{1}{l|}{361.69} & \multicolumn{1}{l|}{\textbf{660.16}} & \multicolumn{1}{l|}{273.67} & \multicolumn{1}{l|}{\textbf{1,128.21}} & \multicolumn{1}{l|}{414.26} & \multicolumn{1}{l|}{\textbf{865.29}} & \multicolumn{1}{l|}{213.22} & \multicolumn{1}{l|}{\textbf{957.25}}\\ \hline

\multicolumn{1}{|l|}{Keras 3 (PT)} & \multicolumn{1}{l|}{1,388.87} & \multicolumn{1}{l|}{2,973.64} & \multicolumn{1}{l|}{525.15} & \multicolumn{1}{l|}{$7,952.67^{*}$} & \multicolumn{1}{l|}{1320.441} & \multicolumn{1}{l|}{3869.72}  & \multicolumn{1}{l|}{452.12} & \multicolumn{1}{l|}{$10932.59^{*}$}\\ \hline
\multicolumn{1}{|l|}{Keras 3 (best)}  & \multicolumn{1}{l|}{355.25} & \multicolumn{1}{l|}{660.16} & \multicolumn{1}{l|}{232.52} & \multicolumn{1}{l|}{1,128.21} & \multicolumn{1}{l|}{404.17} & \multicolumn{1}{l|}{865.29}  & \multicolumn{1}{l|}{185.92} & \multicolumn{1}{l|}{957.25}\\ \hline

\end{tabular}

\caption{Average time taken (in ms/step) per training or inference step across different models, namely SAM~\citep{Kirillov_2023_ICCV}, Gemma~\citep{gemmateam2024gemma}, BERT~\citep{devlin-etal-2019-bert} and Mistral~\citep{jiang2023mistral}.\\
\scriptsize{* LLM inference with the PyTorch backend is abnormally slow at this time because KerasNLP
uses static sequence padding. This will be addressed soon.}}
\label{tab:performance}
\end{table}

\section{Performance}

Framework performance depends on the specific model. Keras 3 offers flexibility by letting users select the fastest framework for their task. Picking the fastest backend for a given model consistently outperforms Keras 2 as seen in Table 1. All benchmarks are done with a single NVIDIA A100 GPU with 40GB of GPU memory on a Google Cloud Compute Engine of machine type a2-highgpu-1g with 12 vCPUs and 85GB host memory.

For fair comparison, we use the same batch size across frameworks if it is the same model and task (fit or predict). However, for different models and tasks, due to their different sizes and architectures, we use different batch sizes to avoid either running out of memory (too large) or under GPU utilization (too small). We also used the same batch size for Gemma and Mistral since they are the same model type with similar number of parameters. (see Table~\ref{tab:performance}).

\section{Related Work}

A library with clear parallels to KerasNLP and KerasCV is the HuggingFace Transformers library~\citep{wolf2020transformers}. Both libraries offer access to pretrained model checkpoints for a number of widely-used transformer architectures.

The Transformers library is built with a ``repeat yourself" approach\footnote{\raggedright{\url{https://huggingface.co/blog/transformers-design-philosophy}}}. KerasNLP, in contrast, is built with a layered approach, with an explicit goal of allowing the re-implementation of any large language model in a relatively small amount of code. We believe there are strengths and weaknesses to both of these approaches.

\section{Future Work}

Future work will focus on extending the project's capabilities. We plan to expand our multimodal model offerings, supporting a wider range of applications. Additionally, we will optimize integrations with backend-specific large model serving solutions, ensuring seamless deployment and scalability.

\section{Conclusions}

KerasCV and KerasNLP are new toolboxes offering both modular components for rapid prototyping of new models, as well as standard pretrained backbones and task models for many computer vision and natural language processing workflows. They can be leveraged by users of either JAX~\citep{jax2018github}, TensorFlow~\citep{tensorflow2015-whitepaper}, or PyTorch~\citep{NEURIPS2019_9015}. Thanks to backend optionality and XLA~\citep{50530} compilation, KerasCV and KerasNLP deliver state-of-the-art training and inference performance. KerasCV\footnote{\url{https://keras.io/guides/keras_cv/}} and KerasNLP\footnote{\url{https://keras.io/guides/keras_nlp/}} offer extensive user guides, available at Keras.io.


\acks{We thank all contributors to Keras \citep{chollet2015keras}, KerasCV \citep{kerascv}, KerasNLP \citep{kerasnlp}, TensorFlow \citep{tensorflow2015-whitepaper},
TensorFlow Text, TensorFlow Data, and the XLA \citep{50530}
compiler, all of which are crucial to the functionality provided in KerasCV and KerasNLP.}


\newpage

\appendix
\section{Preprocessing Layers}
KerasCV provides a comprehensive suite of preprocessing layers that empower users to construct state-of-the-art, industry-grade data augmentation pipelines for image classification, object detection, image segmentation and image generation tasks. These layers implement a wide range of commonly used data augmentation techniques, enabling users to effortlessly enhance the robustness and generalizability of their models. By using preprocessing layers, users can ensure that their models are trained on data that is representative of the data that they will encounter at inference time. KerasCV offers 38 data augmentation layers. These layers implement a wide range of commonly used data augmentation techniques, enabling users to effortlessly manipulate image data in a variety of ways and handle all types of labels out-of-the-box (e.g. class labels, box labels, mask labels).

TF Data is a TensorFlow API for building input pipelines. Input pipelines are responsible for loading data from disk, preprocessing it, and batching it. TF Data provides a number of features that make it a powerful tool for preprocessing data for machine learning, such as:
\begin{itemize}
  \item \textbf{Dataset APIs}: for loading data from a variety of sources, such as CSV files, \verb|TFRecords|, and images.
  \item \textbf{Preprocessing functions}: for performing common preprocessing tasks, such as decoding images, resizing images, and normalizing images.
  \item \textbf{Batching functions}: for grouping data into batches.
  \item \textbf{Prefetching and caching}: for improving the performance of input pipelines.
\end{itemize}

\begin{lstlisting}[language=Python]
# Apply grayscale preprocessing to input
(images, labels), _ = keras.datasets.cifar10.load_data()
to_grayscale = keras_cv.layers.preprocessing.Grayscale()
augmented_images = to_grayscale(images)
\end{lstlisting}

\section{Preset API}
The presets API provides a convenient way to create state-of-the-art CV and NLP models. Presets are pre-configured models that have been trained on a specific dataset and can be used for a specific task.

To use the presets API, one simply needs to import the \verb|keras_cv.models| or \verb|keras_nlp.models| module and then call the \verb|from_preset()| method on the desired model class. The presets API provides a number of advantages over creating models from scratch. First, presets are pretrained on a large dataset, which means that they can achieve high accuracy on a variety of tasks. Second, presets are pre-configured, which means that users do not need to worry about setting hyperparameters. Third, presets are easy to use, which means that users can get started with them quickly.

\begin{lstlisting}[language=Python]
# Load architecture and weights from preset
model = keras_cv.models.RetinaNet.from_preset(
    "resnet50_imagenet",
)

# Load randomly initialized model from preset architecture with weights
model = keras_cv.models.RetinaNet.from_preset(
    "resnet50_imagenet",
    load_weights=False,
    )
\end{lstlisting}

\section{Backbone API}
Both KerasCV and KerasNLP offer a Backbone API. Backbones can be thought of as the central architecture of a model, without the final output layer. This allows users to leverage powerful pretrained backbones (often trained on vast datasets) as the starting point for their own customized models. The pretrained backbones within KerasCV and KerasNLP offer more than just a starting point, they are also fine-tunable. Several examples of how to do this can be seen on the Keras.io webpage~\citep{chollet2015keras}.

\section{Task Models}
KerasCV and KerasNLP provide a number of task models that are designed for specific tasks. These task models are built on top of the KerasCV and KerasNLP modeling layers and provide a high-level of performance. These models are ready for use in applications, but can be further fine-tuned if desired. Some examples of available task models include image classification, object detection, semantic segmentation, image generation, text generation, text classification, and question answering.

\paragraph{Pretrained Task Models} Pretrained task models can be used by using presets trained on different datasets. This allows users to quickly and easily get started with deep learning without having to train a model from scratch. For example, KerasCV provides a number of presets for image classification models that have been trained on different datasets, such as ImageNet, COCO, and Pascal VOC. These presets can be used to create models that can achieve state-of-the-art results on a variety of image classification tasks. To use a pretrained task model with a preset, one simply needs to import the \verb|keras_cv.models| or \verb|keras_nlp.models| module and then call the \verb|from_preset()| method on the desired model class.

\paragraph{Specifying a Backbone in a Task Model} It is possible to specify a backbone for task models. This is done by passing the backbone argument to the \verb|from_preset()| method. By specifying a different backbone, users can change the features that are extracted. This can be useful if one wants to improve the performance of the model on a specific task. User can also specify their own custom backbones. To do this, one simply need to create a subclass of the \verb|models.Backbone| class.

\paragraph{Fine-Tuning a Task Model}
Fine-tuning a task model is the process of adapting a pretrained model to a specific task. This is done by training the model on a dataset of labeled data for the specific task.

\begin{lstlisting}[language=Python]
semantic_segmentation_model = keras_cv.models.DeepLabV3Plus.from_preset(
    "resnet50_v2_imagenet", num_classes=NUM_CLASSES
)
\end{lstlisting}

\vskip 0.2in
\bibliography{citations}

\end{document}